\documentclass[journal]{IEEEtran}
\ifCLASSINFOpdf
\else
\fi
\newcommand{\tick}{\ding{52}}
\newcommand{\cross}{\ding{55}}
\usepackage{graphicx}
\usepackage{longtable}
\usepackage{setspace}
\usepackage{float}
\usepackage{booktabs}
\usepackage{multirow}
\usepackage{lscape}
\usepackage{pifont}
\usepackage{colortbl}
\usepackage[utf8x]{inputenc}
\usepackage{amsmath}
\usepackage[normalem]{ulem}
\usepackage{xcolor}

\newcommand{\PreserveBackslash}[1]{\let\temp=\\#1\let\\=\temp}
\newcolumntype{M}[1]{>{\PreserveBackslash\raggedright}m{#1}}

\begin{document}
\title{Time Perception: A Review on Psychological, Computational and Robotic Models}

\author{Hamit Basgol,
        Inci Ayhan,
        and Emre Ugur % <-this % stops a space
\thanks{H. Basgol is an MA Student in Department of Cognitive Science at Bogazici University, Turkey 
(e-mail: hamitbasgol@gmail.com).}% <-this % stops a space
\thanks{Inci Ayhan is with Department of Psychology in Bogazici University, Istanbul, Turkey (e-mail: inci.ayhan@boun.edu.tr). }% <-this % stops a space
\thanks{Emre Ugur is with Department of Computer Engineering in Bogazici University, Istanbul, Turkey (e-mail: emre.ugur@boun.edu.tr).}}


% \markboth{Transactions in Cognitive and Developmental Systems ,~Vol.~14, No.~8, August~2020}%
% {Shell \MakeLowercase{\textit{et al.}}: Bare Demo of IEEEtran.cls for IEEE Journals}

\markboth{Time Perception}%
{Shell \MakeLowercase{\textit{et al.}}: Bare Demo of IEEEtran.cls for IEEE Journals}




\maketitle

\begin{abstract}
Animals exploit time to survive in the world. Temporal information is required for higher-level cognitive abilities such as planning, decision making, communication, and effective cooperation. Since time is an inseparable part of cognition, there is a growing interest in the artificial intelligence approach to subjective time, which has a possibility of advancing the field. The current survey study aims to provide researchers with an interdisciplinary perspective on time perception. Firstly, we introduce a brief background from the psychology and neuroscience literature, covering the characteristics and models of time perception and related abilities. Secondly, we summarize the emergent computational and robotic models of time perception. A general overview to the literature reveals that a substantial amount of timing models are based on a dedicated time processing like the emergence of a clock-like mechanism from the neural network dynamics and reveal a relationship between the embodiment and time perception. We also notice that most models of timing are developed for either sensory timing (i.e. ability to assess an interval) or motor timing (i.e. ability to reproduce an interval). The number of timing models capable of retrospective timing, which is the ability to track time without paying attention, is insufficient. In this light, we discuss the possible research directions to promote interdisciplinary collaboration in the field of time perception.
\end{abstract}

% % Note that keywords are not normally used for peerreview papers.
% \begin{IEEEkeywords}
% IEEE, IEEEtran, journal, \LaTeX, paper, template.
% \end{IEEEkeywords}

% \fi
\IEEEpeerreviewmaketitle

\section{Introduction} \label{SEC:INTR}

Time, according to Kant \cite{burnham2007kant}, along with space, is a main parameter constituting the possibility of knowledge. Since time conveys information regarding the current and future state of the environment, biological systems can organize their functions, behaviors, and cognitive abilities according to temporal information \cite{mihailovic2017time}. {\color{black}Animals can quantify the elapsed time with or without the external stimuli (sensing), store this information in the memory (memorizing), and anticipate future states of the environment (predicting) \cite{merchant2020estimating}.} It was shown that a wide range of animals is capable of time-place learning, which is the ability to associate the subjective place and time for the avoidance from predators, localization of resources, and gaining a survival advantage \cite{mulder2013circadian}. Moreover, it was found that vertebrates having smaller bodies and higher metabolic rates perceive time passing slower in comparison to ones having larger bodies and lower metabolic rates because perceiving in higher temporal resolution poses an energetic cost \cite{healy2013metabolic}. The difference in subjective time perception as a function of body size and metabolic rate affects the amount of saved energy, and accordingly, provides a survival advantage to animals \cite{healy2013metabolic}. These findings are not surprising because we know that animals navigate not only in space but also in time in order to show robust and adaptive behaviors. Thus, it can be concluded that animals do not live in a three-dimensional world but rather in a four-dimensional one, involving time.

\begin{figure}[t]
\centering
  \includegraphics[width=\columnwidth]{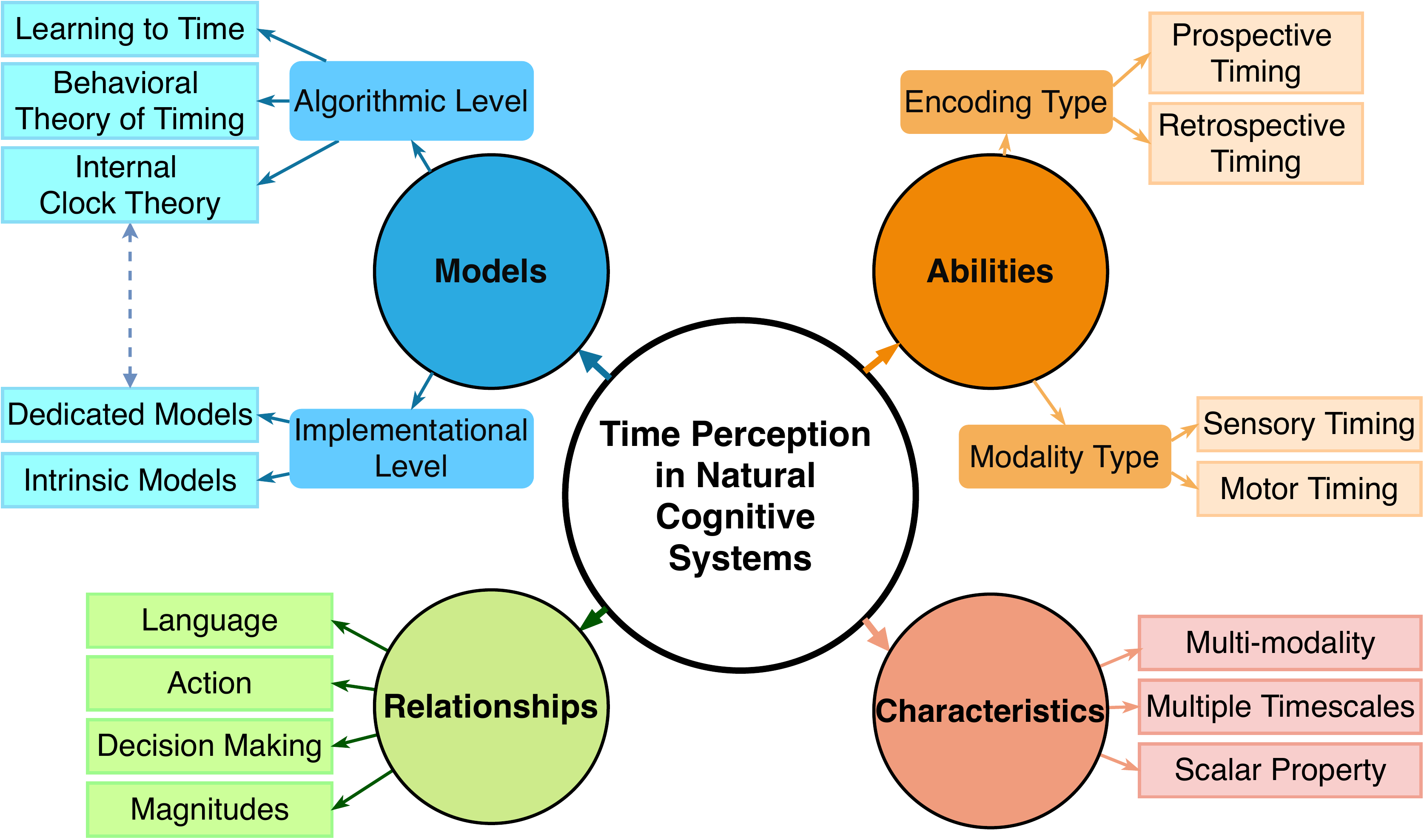}
\caption{Mind map for natural cognitive systems of time perception}
\label{fig:naturalcognitivesystems}
\end{figure}

Animals are robust and adaptive biological systems. From this perspective, according to Pfeifer et al. \cite{pfeifer2007self}, understanding the mechanisms underlying the biological processes might be a source of inspiration for developing robust and adaptive systems to the environmental changes and perturbations. In this context, embodied artificial intelligence takes inspiration from biological systems. Studying how biological systems acquire temporal information and how they use it to achieve sensory, motor, and cognitive processes is an essential topic for further research in embodied artificial intelligence, cognitive robotics and computational psychology. This type of research might reveal two positive outcomes: The first one is the exploitation of temporal information by artificial intelligence systems. As emphasized by Maniadakis et al. \cite{maniadakis2011temporal, maniadakis2011time}, the use of temporal information in artificial systems has been limited, although it is necessary to develop intelligent systems that efficiently interact with their environments. For example, in human-robot interaction scenarios, intentions of agents' behaviors, and their causes are not directly observable. Think about a shared work-space where a robot should collaborate with different human partners, each with different working styles. Even though it is the very task, some people might operate in a rush because of perceived time pressure or a personality trait. The differentiation between these two possibilities requires the acquisition of fine temporal dynamics of behaviors. Therefore, the value of this type of knowledge is tremendous for autonomous systems. Timing is also crucial for user satisfaction. {\color{black} A recent report reveals that the temporal precision in the timing of robots' behaviors affects user experience ratings more than the spatial precision \cite{koene2014relative}.} On the other hand, time perception tasks in animal and human timing can be utilized to figure out whether algorithms developed by artificial intelligence researchers can exploit temporal information and, if they can, how these algorithms achieve this ability \cite{deverett2019interval}.

\begin{figure}[t]
\centering
  \includegraphics[width=\columnwidth]{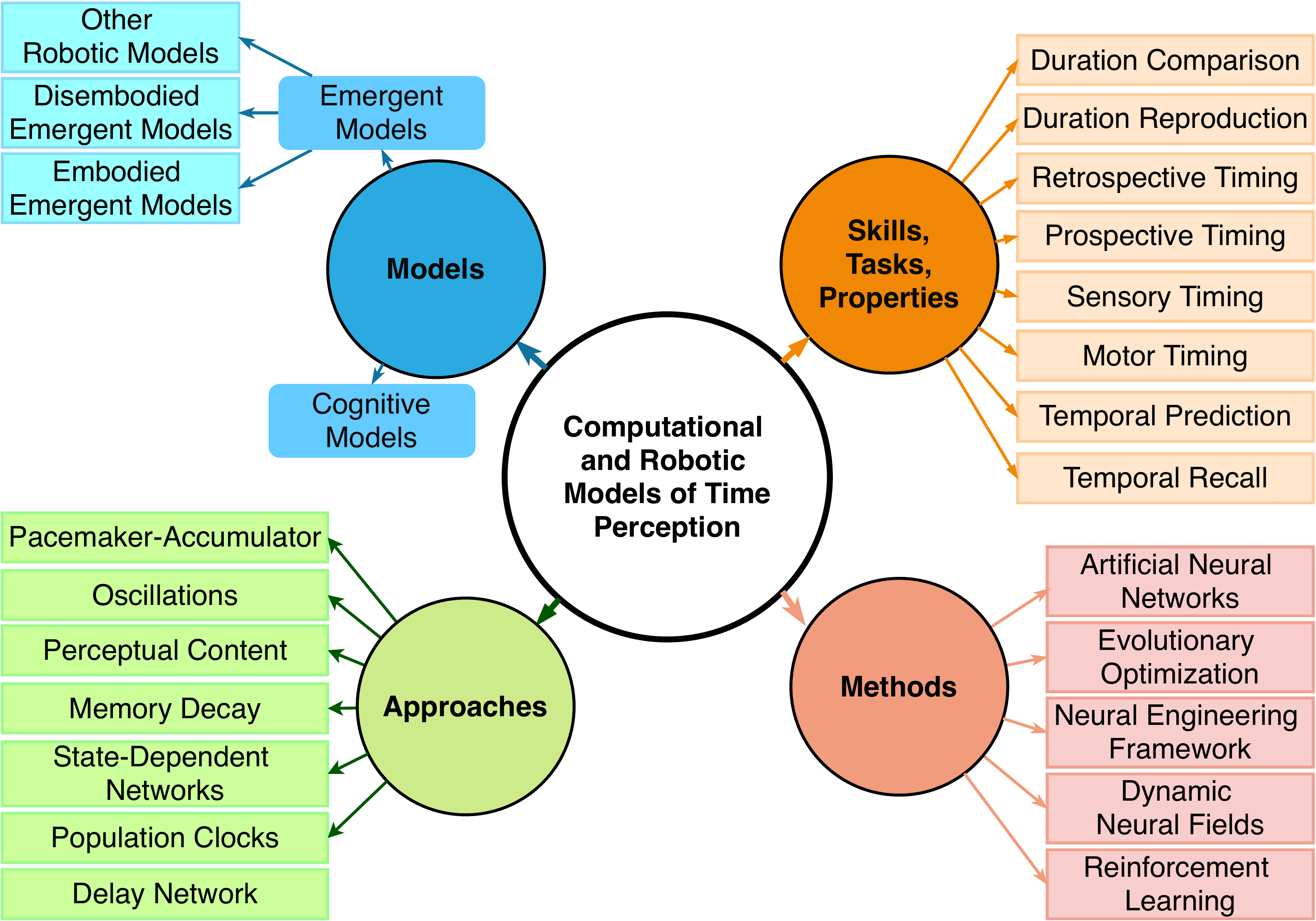}
\caption{Mind map for computational and robotic models of time perception}
\label{fig:compcognitivesystems}
\end{figure}

The second outcome of computationally studying how biological systems use temporal information is obtaining plausible hypotheses and remarkable insights about the time perception mechanisms in biological systems \cite{addyman2016computational, deverett2019interval, hardy2016neurocomputational, maniadakis2011time}. {\color{black}In this manner, researchers may develop models for specific timing tasks devised in animal and human timing literature to shed light on possible timing mechanisms.} In fact, there is a growing interest in developing computational and robotic models that can use temporal information \cite{addyman2016computational, deverett2019interval, duran2017learning, hourdakis2018robust, koskinopoulou2018learning, maniadakis2011time, maniadakis2012experiencing, maniadakis2014time, maniadakis2015integrated, roseboom2019activity}. A review study integrating knowledge about time perception across all disciplines, including psychology, cognitive science, neuroscience, and artificial intelligence, on the other hand, seems to be lacking. {\color{black}Thus, here, in order to achieve two outcomes that have been described so far, we report recent findings in the literature and discuss possible research directions to promote interdisciplinary collaboration in the future.}

This study is composed of three sections: The first section briefly discusses the necessary concepts regarding the use and processing of temporal information in natural cognitive systems (Fig.~\ref{fig:naturalcognitivesystems}). The second section describes the computational and robotic models of time perception. We categorize them into two groups, namely cognitive and emergent models, and limit our discussion to the emergent models (Fig.~\ref{fig:compcognitivesystems}). Finally, we provide a general discussion on the current status of the literature and present a set of possible research questions.

% It seems that we learn behavioral patterns of things and temporal relationships between behavioral units. A crucial task for a robot would be forming a collaboration with. It can be said that some people are slow in their behavior; others are fast. Carrying out complex actions with humans in a shared work-space might necessitate learning the fine temporal dynamics of behaviors of each human partner. In terms of

\section{Time Perception in Natural Cognitive Systems} \label{tpncs}

% Thanks to a basic intuition and the need for clarity, we categorized cognitive systems into two kinds as natural and artificial. By natural, we mean biological cognitive systems; by artificial, we refer to cognitive systems being a product of engineering efforts. 

{\color{black}{Here we briefly discuss time perception in natural cognitive systems. In the first subsection, we examine the distinguishing characteristics of time perception in animals, including humans. We emphasize the connection between time and other cognitive abilities, the maturation of time perception throughout the development, and time perception tasks and abilities. In the next subsection, we elaborate on the classical time perception models explaining how animals use temporal information. These topics will help us to refine our discussion about computational and robotic models of time perception.}}

\subsection{Characteristics of Time Perception}

\subsubsection{Multi-modality of time perception} Time perception has several distinguishing characteristics. One of them is that subjective time perception is formed by the interaction between different sensory modalities \cite{bausenhart2014multimodal, vroomen2010perception}. For example, think about a person talking on TV. Physically, mouth movements and language are out of sync; however, we perceive them as if they happen simultaneously. This phenomenon is called the \emph{temporal ventriloquism} \cite{bausenhart2014multimodal} and shows the fact that time perception is \emph{multi-modal}.

% In the spatial domain, visual information biases the perceived location of auditory stimulus, which is called \emph{spatial ventriloquism}. It can be seen while watching TV where the sound comes from speakers rather than people's mouth. 

\subsubsection{Timescales of time perception} Animals can use temporal information in different timescales, which can be divided to four, namely \emph{microsecond timing}, \emph{millisecond timing}, \emph{second timing}, and \emph{circadian timing}, each of which contributes to different abilities in organisms' lives \cite{buhusi2005makes}. For example, timing up to milliseconds was shown to be crucial for producing speech \cite{schirmer2004timing} and motor control \cite{sober2018millisecond}, whereas timing between seconds to minutes is essential for working memory maintenance \cite{brody2003timing} and the production of action sequences \cite{bortoletto2011motor}. Depending on the day-night cycle, circadian timing regulates the sleep-wake cycle and metabolism \cite{buhusi2005makes, czeisler1999stability}. In this paper, we restrict our discussion to milliseconds-to-seconds and seconds-to-hours. 

\begin{figure*}[!t]
\center
    \includegraphics[width=\textwidth]{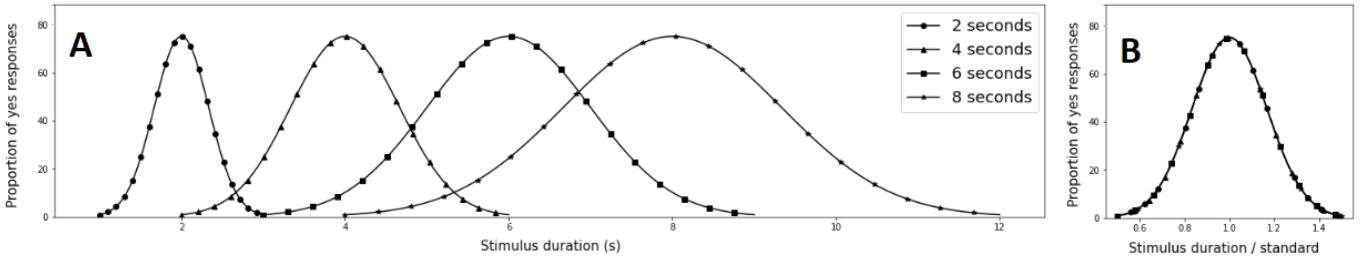}
    \caption{The figure shows a depiction of scalar property. In a \emph{temporal generalization task}, animals receive a standard and a target stimulus and are trained to press yes when the standard stimulus is the same as the target stimulus. In the figure, different experimental groups (standard stimulus with a duration of 2, 4, 6, and 8 seconds) are given in the x-axis. (A) The maximum proportion of yes responses is converged to the real duration and the variance of the proportion of yes responses increases as the duration to be estimated increases. (B) Moreover, the increase in variance is linearly proportional to the estimated duration. This is the scalar property of time perception. {\color{black}The figure is adapted from \cite{wearden1997scalar}.}}
\label{fig:scalar_property}
\end{figure*}

\subsubsection{The scalar property of time perception} The discrimination of two perceptual quantities depends on the ratio between them, which is called the \emph{Weber's law}. Weber's law is seen in different domains such as number, length, and duration and reveals itself as a \emph{scalar property} in duration discrimination (see Fig.~\ref{fig:scalar_property}) \cite{matell2004cortico}. The scalar property defines a strict mathematical relationship between the estimations (target duration) and the interval being estimated (standard duration). It refers to the fact that as the estimated duration gets longer, the deviation of estimations from the standard duration also increases \emph{linearly}. For the scalar property, ``the standard deviations of time estimates grow as a constant fraction of the mean,'' \cite[p. 218]{ferrara1997changing} meaning that the coefficient of variation statistic (standard deviation/mean or CV) remains constant. This property has been observed in duration estimation performances of several animals such as rats and pigeons \cite{buhusi2009interval, lejeune2006scalar, malapani2002scalar}. For very short ($<$ 100 ms) and long durations ($>$ 100 s) and in particular for the challenging tasks, on the other hand, deviations from the scalar property have also been observed  \cite{lejeune2006scalar}. For instance, Ferrara et al. \cite{ferrara1997changing} conducted a study with two conditions in which participants were to detect whether the target duration is the same as the standard duration (see Fig.~\ref{fig:scalar_property}). For the easy condition, target durations were set to be around 600 ms with 150 ms increments, whereas for the hard condition, with smaller increments (75 ms). Surprisingly, the group in the hard condition was more sensitive to the difference in the two stimulus durations than that in the easy condition. This result suggests that the sensitivity of the timing system varies according to the discrimination difficulty of two temporal intervals. This violates the scalar property, predicting the same sensitivity rate for all durations.

Despite its distinguishing characteristics, the ability to time is not isolated. In fact, it scaffolds other perceptual, motor, and cognitive processes. In the next subsection, we will try to shed light on the relationship between time perception and other cognitive abilities.

\subsubsection{Time perception and related cognitive abilities}

Time has a vital role in performing actions. It was found that the duration between action execution and the expected sensory input affects the sense of agency \cite{stetson2006motor}, which in turn affects the perceived duration in between \cite{haggard2002voluntary, moore2012intentional}. The former was observed in the \emph{sensory-motor temporal calibration paradigm}, while the latter evidence was observed in the \emph{intentional binding paradigm}. In the sensory-motor temporal calibration paradigm, researchers put an artificial lag between a button press (action) and a beep sound (effect). After the training, the lag is removed and participants perceive as if the effect occurred before the action \cite{stetson2006motor}. An opposite effect is seen in the \emph{intentional binding paradigm} such that in experimental setups, where participants think that they are the causal agent for the sensory effect, the duration between the button pressing (action) and the beep sound (effect) is perceived to be compressed \cite{haggard2002voluntary, moore2012intentional}. These two paradigms show that time plays a role in forming the sense of agency and connecting the action (cause) and sensory effect into one another. 

In addition to the binding of an action and its effect, time is also essential for decision making \cite{klapproth2008time}. For example, in the classical tasks shown in Fig.~\ref{fig:temporal_abilities}A and Fig.~\ref{fig:temporal_abilities}B, making a decision requires the ability to estimate time. Time is also essential for decision making in a real-world context, which can be seen in Fig.~\ref{fig:temporal_abilities}E. In their seminal work, Leclerc et al. \cite{leclerc1995waiting} showed that people tend to choose the events with known durations while making a decision amongst a set of options. Additionally, time determines the value of outcomes. In fact, for an agent, immediate and delayed outcomes do not have the same value, which is called the \emph{temporal discounting} \cite{critchfield2001temporal}. It is believed that temporal discounting is a personality trait affecting people's ability to make long-term plans \cite{simons2004placing}. 

Investigating the connection between language and time perception, Wearden et al. \cite{wearden2008perception} focused on speech control and metaphor comprehension. The former is an example of how time perception affects language understanding and use and the latter is an example of how language affects perceiving time. The research suggested that duration discrimination problems in speech result in speech perception and production problems \cite{tallal2004improving}; children having poor reading abilities also have poor temporal judgment capabilities \cite{may1988temporal} and training for temporal discrimination improves phonetic identification \cite{szymaszek2018training}. It is also claimed that language determines how we perceive time. Boroditsky \cite{boroditsky2001does} observed that English people perceive time as if it flows horizontally, whereas Mandarin people think that it flows vertically. Moreover, Hendricks and Boroditsky \cite{hendricks2017new} showed that learning a new metaphor to talk about time leads people to form its non-linguistic representations.

\subsubsection{Time perception and magnitude perception} 

A stimulus has measurable properties, namely magnitudes, such as its volume in space, number and duration. A substantial amount of research shows that the magnitude perception skills are not isolated from one another. For example, Brannon et al. \cite{brannon2006development} found that infants who are at their six months of age show the same sensitivity to number, time, and area in a discrimination task. Xuan et al. \cite{xuan2007larger} showed that the error in temporal judgment is affected by other magnitudes such as number, size, and luminance. The relationship between magnitudes leads researchers to think that there can be a common magnitude representation system in the brain. This idea was theorized by Walsh \cite{bueti2009parietal, walsh2003theory} as \emph{a theory of magnitude (ATOM)}. According to the ATOM, time, space and number are sensory-motor decision variables that are used for action execution. For this reason, they are processed in a common magnitude system located in the inferior parietal cortex. Coming from birth, this system is hardwired by the evolutionary process \cite{walsh2003theory, winter2015magnitudes}. ATOM proposes a map between magnitudes from birth, while other researchers suggest that this map might be learnt through experience. Cantlon \cite{cantlon2012math} listed the possible explanations as to how the mapping between different magnitudes is established. 

Timing is not isolated from the other cognitive abilities and magnitude types. Thus, many models and theories are developed to achieve an integration \cite{taatgen2007integrated, walsh2003theory}. At the same time, timing is not stable or static from birth, either, but rather, it maturates and changes throughout the development. In the next subsection, we will give a brief outline on the development of time perception.

\subsubsection{The development of time perception}

\begin{figure*}[t!]
\centering
\includegraphics[width=\textwidth]{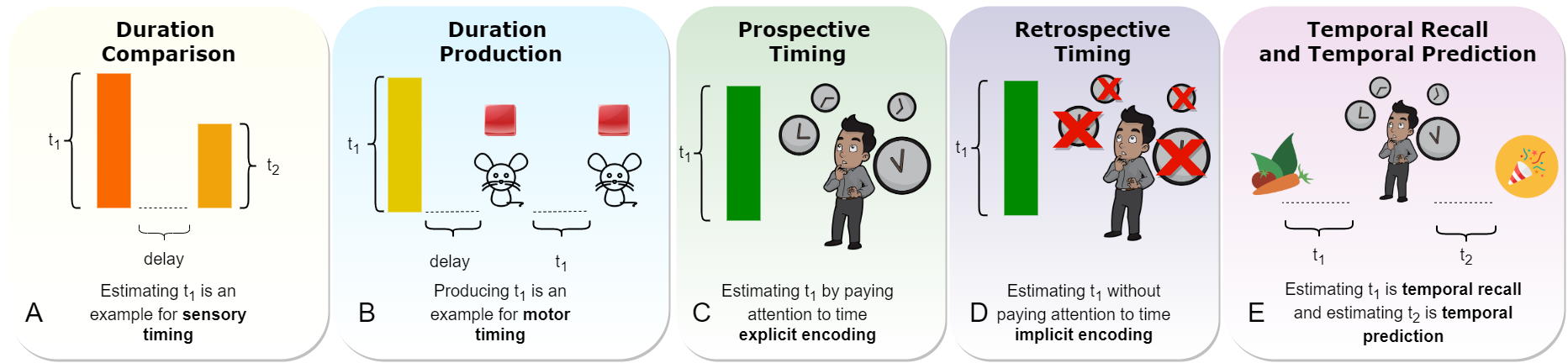}
\caption{The figure shows interval timing abilities and related tasks. (A) In a classical duration comparison task, the agent is asked to decide which stimulus is longer or shorter (t\textsubscript{1} and t\textsubscript{2}). The task requires estimating the duration of t\textsubscript{1} and t\textsubscript{2}. Duration estimation for one sensory stimulus is called sensory timing. (B) In a classical duration reproduction task, an agent is given a target duration that should be produced by marking the start and the end of the event by pressing a button. Producing t\textsubscript{1} requires motor timing ability. (C) One can estimate the duration of an event by paying attention or (D) without knowing. E) In addition to these abilities, one can estimate when an event occurred or would occur.}
\label{fig:temporal_abilities}
\end{figure*}

In the course of development, human babies' timing abilities show substantial changes \cite{mccormack2015development}. Despite these changes, though, in the early years of their lives, they still hold remarkable skills. For example, the evidence demonstrates that infants form temporal predictions \cite{colombo2002infant} and are sensitive to the interval between two stimuli \cite{brannon2004timing, brannon2008electrophysiological}. Further research reveals that the infants' ability to discriminate different durations develops throughout the course of their development. It was observed that 3-month-old infants can discriminate durations with 1:3 \cite{gava2012discrimination}, 5- to 6-month-old infants with 1:2 \cite{vanmarle2006six}, 10 month-old infants with 2:3 ratio \cite{brannon2007temporal} (t\textsubscript{1}:t\textsubscript{2} in Fig.~\ref{fig:temporal_abilities}A). That is, although infants can process temporal information from birth, they maturate to distinguish more complex proportions throughout the childhood \cite{allman2012developmental}.

Up to here, we provided a brief outline on time perception. In the following subsections, we elaborate on the basic timing abilities and temporal information processing models. 

\subsubsection{Time perception tasks and abilities} 

\emph{Duration} is a feature of sensory stimulus having an onset and an offset. Ability to represent duration in seconds-to-hours \cite{oprisan2014all} and milliseconds-to-seconds \cite{paton2018neural} is considered as \emph{interval timing}. Studies conducted for investigating animals' interval timing abilities revealed important tasks (Fig.~\ref{fig:temporal_abilities}).

Interval timing tasks can be grouped into two major classes according to the use of temporal information. While tasks requiring the estimation of sensory stimuli are named as \emph{sensory timing} tasks, tasks requiring the regeneration of duration information are called \emph{motor timing} tasks. That is, sensory timing is about how much time has passed, while motor timing is about when or how long a behavior will be shown. For motor timing, temporal information should be reproduced by motor commands \cite{buonomano2011population}. In Fig.\ref{fig:temporal_abilities}A\&B, basic tasks for sensory and motor timing are shown. 

A further categorization between timing tasks can be made with respect to the type of encoding. Animals can encode temporal relationships of environmental dynamics unconsciously. This is called implicit encoding of temporal information and assessed using a \emph{retrospective timing} task in which subjects are not pre-informed that they would be asked to estimate a duration \cite{block2018prospective, grondin2010timing} (Fig.~\ref{fig:temporal_abilities}D). If subjects know that they would be asked to estimate a duration, the task is a \emph{prospective timing} task (Fig.~\ref{fig:temporal_abilities}C) \cite{block2018prospective, grondin2010timing}. {\color{black}{Being engaged in the timing task, cognitive system uses attentional resources in prospective timing. In general, prospective timing performances are less variable than retrospective timing performances and the interval to be estimated is perceived longer in prospective timing \cite{block2018prospective}.}} For example, if subjects are asked to guess how long the computer has been open, this is a retrospective timing task because subjects do not deliberately track the duration. Since they implicitly encode it, they should guess by counting on their memory. If subjects are asked to wait and deliberately track how long the computer will be open until it is turned off, this is a prospective timing task because subjects can give their attention to the temporal information throughout the task.

In addition to the encoding of time, whether the estimated duration is in the past or in the future is another conceptual distinction (Fig.~\ref{fig:temporal_abilities}E). For the past, we can define a term called temporal recall (already named as \emph{timing when} by Maniadakis and Trahanias \cite{maniadakis2016and}), which is the ability to estimate when an event occurred. On the other hand, for the future, temporal prediction is the use of temporal dynamics to assess when an event would occur or be completed.

In the later section, we will evaluate computational and robotic models by asking whether the target model mentions explicit or implicit encoding (prospective and retrospective timing), whether it is capable of representing or regenerating the duration (sensory and motor timing), and whether it shows scalar property (Fig.~\ref{fig:temporal_abilities}). Before the computational and robotic models, however, in the next subsection, we will first summarize the temporal information processing models that aim to explain performances in human and animal timing.

\subsection{Temporal Information Processing Models} 

Explaining how animals process temporal information is the central tenet of time perception research. In the literature, two types of models, namely \emph{dedicated} and \emph{intrinsic models}, are the two competing explanations \cite{ivry2008dedicated}. As for the dedicated models, specialized functions contributing to temporal information processing are localized on the same part, whereas different functions are localized on different parts of the brain. On the other hand, \emph{intrinsic models} postulate that temporal information processing does not depend on specific brain regions but is a function of neural populations \cite{ivry2008dedicated}. Dedicated and intrinsic models consider the biological basis of timing; in other words, they are in the implementational level of explanation. Besides the implementational level, these models are influenced by an information processing model in the algorithmic level, namely \emph{internal clock theory}, which suggests that there are specialized processes and representations for timing. Since the theory assumes specialized brain areas for timing, it has a close relationship with dedicated models of time perception \cite{church1984properties, meck1984attentional}. Next, we will investigate the relationship between the internal clock theory and the dedicated models of time perception.

\subsubsection{Internal clock theory and dedicated models} \label{int_clock_ded_model}
\begin{figure}[!b]
\centering
  \includegraphics[width=\columnwidth]{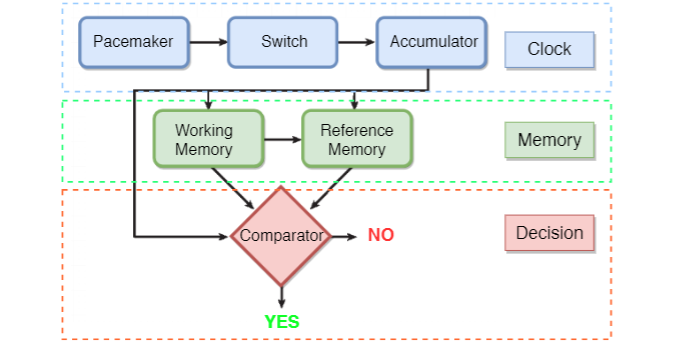}
\caption{The information processing model of the internal clock theory}
\label{fig:internal_clock}
\end{figure}

According to the internal clock theory, a mechanism resembling a clock turns physical time into the subjective experience of time. This theory was put forward as a result of the studies conducted by Treisman \cite{treisman1963temporal} in psychophysics and Gibbon et al. \cite{gibbon1977scalar, gibbon1984scalar} in animal learning. An internal clock is formed with \emph{clock, memory and decision phases} \cite{church1984properties} (Fig.~\ref{fig:internal_clock}). In the clock phase, \emph{pacemaker} module generates rhythmic pulses and sends them to an \emph{accumulator} through a \emph{switch} which determines the frequency of passing pulses. In the memory phase, rhythmic pulses generated in the clock phase are sent to the working and reference memories. While working memory stores the current amount of pulses generated by the pacemaker, reference memory stores the earlier amount of pulses that have been learnt per unit of time. In the decision phase, pulses in the working and reference memories are compared to decide whether they correspond to the same time interval \cite{allman2014properties}. The internal clock theory offers an explanation about how animals learn a duration in a fixed-time interval operant conditioning procedure (FI) \cite{skinner1990behavior}, where an animal learns to press a button at certain temporal intervals to receive a reward. In the initial trials of the procedure, the animal starts pressing the button randomly. As the experiment unfolds, the animal stores the required pulses to press the button in reference memory and presses the button when enough pulses are accumulated.

% {\color{black} Responses of rats trained on either FI-30 or FI-240 seconds are given in Figure \ref{fig:fi_graph}. It can be seen from the graph that distributions of responses of animals in FI-30 and FI-240 are strikingly similar and validate the scalar property of time perception.}

% stating that the likelihood of discrimination does not depend on the absolute difference but depend on the ratio between two quantities. It was suggested that variances in other parts of the clock contributes to this phenomenon \cite{rakitin1998scalar, matell2004cortico}. 

% \begin{figure}[!b]
% \centering
%   \includegraphics[scale=0.25]{FI_graphs.pdf}
% \caption{{\color{black}The figure shows rats' performances on fixed-interval schedules for 30 and 240 seconds. Note that timing is relative to the standard being timed. The figure is adapted from \cite{whitaker2003multiple}.}}
% \label{fig:fi_graph}
% \end{figure}

Specialized functions proposed by the internal clock theory inspire dedicated models, which assume that these functions are realized in the brain. According to the \emph{specialized timing models}, so-called internal clock is hypothesized to be located in one part of the brain, such as cerebellum \cite{ivry2002cerebellum}, basal ganglion \cite{harrington1998temporal}, supplementary motor area \cite{macar2006supplementary} or right prefrontal cortex \cite{lewis2006right}; whereas for the \emph{distributed timing models}, functions of internal clock are distributed in the brain \cite{ivry2008dedicated}. 

There is a substantial amount of work in favor of the internal clock theory. Recall that the theory assumes a pacemaker that generates pulses and an accumulator that stores them (Fig.~\ref{fig:internal_clock}). Treisman et al. \cite{treisman1990internal, treisman1992time} found that repetitive visual and auditory stimuli can affect the frequency of pulses emitted by the pacemaker and therefore change the perceived duration as if it lasted longer. Meck \cite{meck1983selective} showed that the pharmacological manipulations selectively change the performance of memory and decision processes in the internal clock. His work pointed out that the increased dopamine level extends the perceived duration by increasing the number of pulses emitted by the pacemaker. According to Gibbon \cite{gibbon1992ubiquity}, an internal clock which is subject to noise in the encoding and retrieval phases can exhibit scalar property. Further evidence for the theory is related to a property of temporal representations. Since the internal clock is a general time-keeping mechanism, the theory assumes amodal temporal representations, which are transferred from one modality to another without notable performance differences. One evidence for this was found by Keele et al. \cite{keele1985perception}, who showed that timing accuracies of participants estimating time with a finger, foot, or by observation are correlated with each other. In other words, people being successful in sensory timing are also successful at motor timing and vice versa. On the other hand, many influential work question the amodal nature of temporal representations. For example, it was shown that auditory stimuli are experienced longer than visual stimuli, even though they have the same duration \cite{wearden1998sounds}. Similarly, it was suggested that there are different mechanisms for sensory and motor timing \cite{buonomano2011population}. In addition to the modality-dependent temporal representations, subjective time is also multi-modal \cite{bausenhart2014multimodal, chen2013intersensory, vroomen2010perception}. It must be pointed out that the internal clock is a high-level and generic cognitive mechanism. Recent research comes up with considerable challenges with this idea by selectively manipulating the perceived duration of stimuli across visual space \cite{ayhan2009spatial, johnston2006spatially}. This type of manipulation suggests an inherent association between space and time and thus validates a modality-specific timing mechanism in the brain. Moreover, following this research line, \cite{gulhan2019short} questioned whether a time pathway specialized for processing time as a property of visual information exists and found evidence for the relationship between sensory processing and time perception in higher-level motion areas. That is, for brief time intervals, namely milliseconds, there can be a modality-dependent neural pathway for processing time, which connects the early visual system to higher-level cortical areas \cite{ayhan2020action, gulhan2019short}.

Apart from the possibility of modality-dependency and multi-modality of temporal representations, another limitation of the theory is that the assumed internal clock needs a reset point and can only encode the duration of the stimulus explicitly \cite{treisman1963temporal, gibbon1977scalar}. Thus, it gives priority to the prospective estimation of time. Finally, the localization of the internal clock in the brain is still a matter of debate (for candidate brain areas, refer \cite{allman2014properties}). Internal clock theory, despite its limitations, supports an intuitive mechanism.

Intrinsic models \cite{ivry2008dedicated}, on the other hand, aim to explain time perception without depending on a clock-like mechanism. Since intrinsic models propose that neural groups can process temporal information, they are generally immune to the problems faced by the internal clock theory.

\subsubsection{Intrinsic models}

Intrinsic models state that time perception does not depend on specialized brain regions. Theories relying on intrinsic models collaborate intensely with neurocomputational models to investigate the underlying mechanisms of time perception. Relevant to our discussion, these models will be detailed in the following sections.

An intrinsic model which does not have a computational implementation is the \emph{energy readout theory} proposed by \cite{pariyadath2007effect}. Pariyadath and Eagleman explain their theory within the context of an oddball paradigm, in which subjects are presented with a sequence of standard and target stimuli as in Fig.~\ref{fig:scalar_property}. In this paradigm, the target stimulus presented much less frequently than the standard is called the oddball stimulus. Surprisingly, the oddball stimulus is perceived longer than the more frequent stimulus \cite{tse2004attention, ulrich2006perceived}. According to the \emph{energy readout theory}, the magnitude of neural activation codes the duration of stimulus and determines whether the stimulus is perceived as shorter or longer. Since predictability leads to a suppressed neural activation, the subjective duration of the frequent stimulus is shortened.

Intrinsic models consider neural populations as the key actor of time perception and accept that neurons held in the ordinary cognitive tasks might be used for temporal processing \cite{ivry2008dedicated, karmarkar2007timing}. For this reason, they are better at explaining \emph{modality and task-based performance differences} \cite{spencer2009evaluating}. On the other hand, these models cannot explain performance transitions between modalities and are limited to milliseconds \cite{ivry2008dedicated}.

\begin{figure}[t!]
\centering
  \includegraphics[width=\columnwidth]{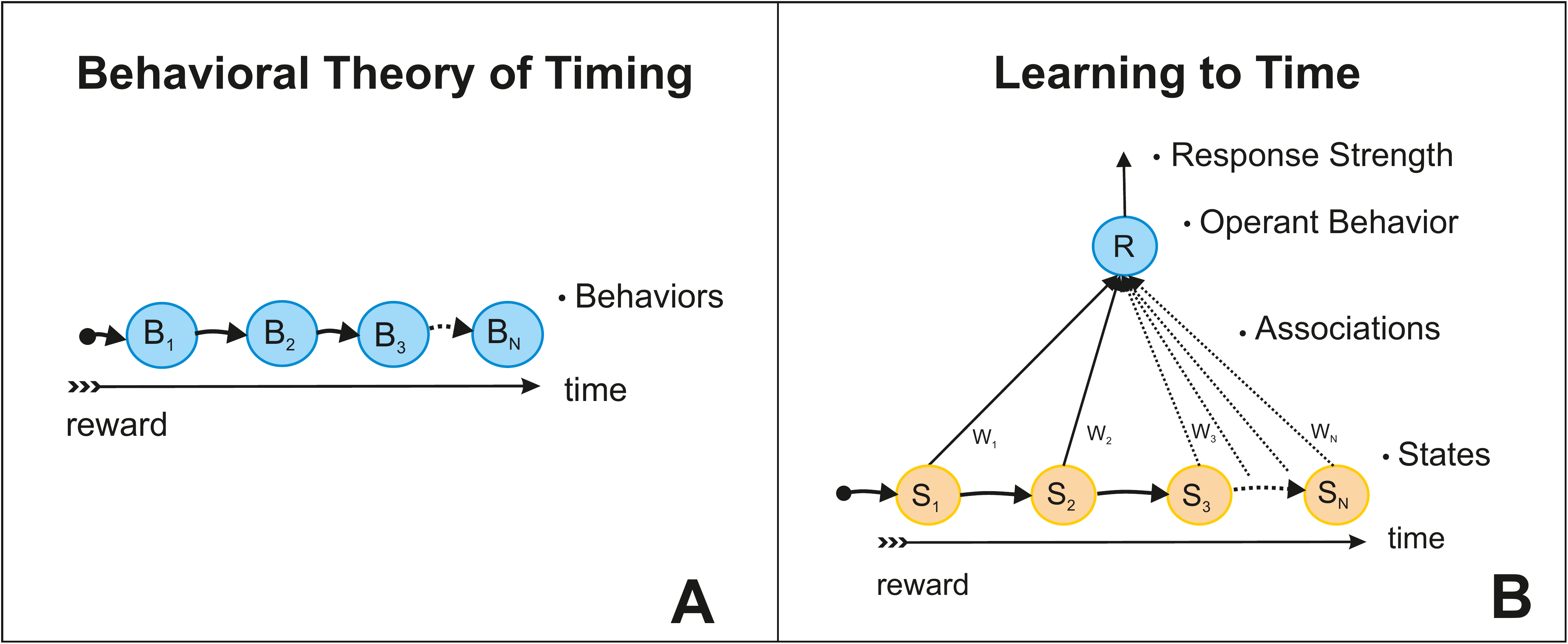}
\caption{{\color{black}The illustration of BeT and LeT models. (A) While BeT assumes that the operant behavior conditioned on adjunctive behaviors leads to timing performance, (B) LeT extends this idea by proposing a learning rule. B is adapted from \cite{wearden2016psychology}.}}
\label{fig:BETLET}
\end{figure}

{\color{black}Apart from the dichotomy between the dedicated and intrinsic models, there are two behavioral models of timing which try to explain animal timing not with an internal, clock-like mechanism but with behaviors learned during the training phase: (a) Behavioral theory of timing and (b) learning to time models. These models ground animal timing ability on successive behaviors activated through time.

\subsubsection{Behavioral theory of timing (BeT) and learning to time (LeT) models}

Behaviorism's central tenet is to explain animal behaviors with their relation to the environment without depending on an internal, cognitive mechanism. Seems to be isolated from the external environment, animal timing ability presents a challenge to behaviorist tradition. 

% To make it concrete, recall the FI schedule where animals do not receive an external cue to distinguish a particular time point from another. Without the help of an external stimulus, how can animals learn to respond at the specific point marked by reward?

Being a creative and ingenious attempt by Killeen and Fetterman \cite{killeen1988behavioral}, BeT pays attention to the adjunctive behaviors that animals start to exhibit after training in FI. Adjunctive behaviors are contingent and not purposefully reinforced by experimenters but emerged during training, such as a mouse drinking water while waiting for the next food delivery. According to BeT, adjunctive behaviors serve as a signal for both other adjunctive behaviors and the operant behavior (i.e. reinforced behavior) (Fig.~\ref{fig:BETLET}). To explain the scalar property, BeT puts forward several assumptions. Firstly, it assumes that the number of adjunctive behaviors is the same regardless of the interval length. Secondly, there is a hypothetical pacemaker whose ticks mark behavioral transitions. Finally, pacemaker speed is determined according to the reinforcement rate. 

% That is, consider the FI-30 and FI-240 schedules; the reinforcement rate of the former is higher than the latter one. Therefore, the hypothesized pacemaker is faster in FI-30 because the number of adjunctive behaviors is constant. Thus, these assumptions make the response shape in FI-30 similar to that in FI-240 (refer Figure \ref{fig:fi_graph}).

Another behavioral model trying to throw light on animal timing is LeT, which was put forward by Machado \cite{machado1997learning} to predict when and how much animals respond in a timing task. LeT assumes that there are successive behavioral states, including ``elicited, induced, adjunctive, interim and terminal classes of behavior'' \cite{machado1997learning} (Fig.~\ref{fig:BETLET}). Throughout trials, associations between behavioral states and the operant response are formed to determine the response rate. Similar to the Hebbian principle explaining the connection formation between neurons, according to LeT, the association between a behavioral state and operant response is strengthened if they occur together. Specific difficulties arise in the generalization of these behavioral theories to human timing ability. For example, they do not account for one-shot human performance in timing tasks. Also, since they are based on behaviors occurring in real-time, they might not be an alternative theory for subsecond intervals \cite{wearden2016psychology}. In comparison to their explanatory difficulties in human timing, these theories can shed light on the timing abilities of reinforcement learning (RL) agents. As opposed to the internal clock theory, behavioral theories on animal timing, namely BeT and LeT, do not presuppose a neural mechanism.

Having provided a brief summary about time perception models and summarized how animals process temporal information, we will now investigate the computational and robotic models of time perception.}

\begin{figure*}[t]
\centering
  \includegraphics[width=\textwidth]{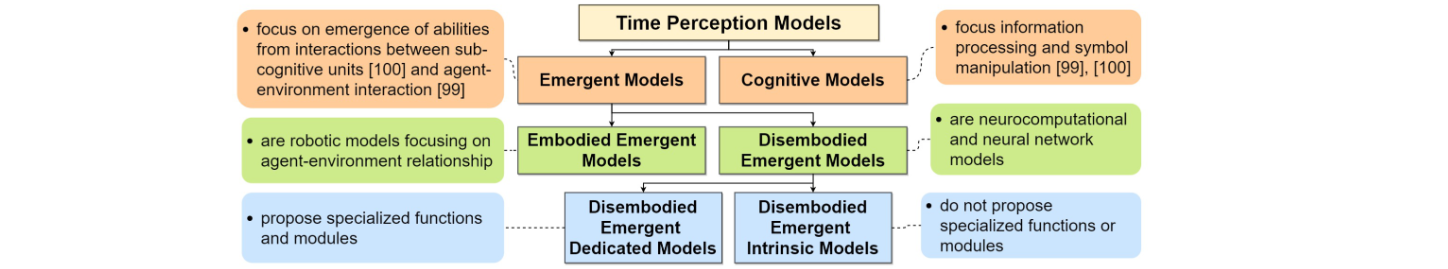}
\caption{Embodied and disembodied models of time perception}
\label{fig:time_perception_models}
\end{figure*}

\section{Computational and Robotic Models of Time Perception} \label{rmtp}

Investigating how animals process temporal information and mimicking the same principles by computational and robotic models enables us to develop robust and adaptive systems. Additionally, time perception tasks can be utilized for evaluating the capabilities of computational agents. It is important to note that the relationship is not one-sided. Investigations with computational agents might reveal possible hypotheses and significant insights into how animals use temporal information in the environment.

According to Vernon et al. \cite{vernon2007survey}, computational models can be classified into two groups: cognitive and emergent models. While \emph{cognitive models} focus on the information processing and symbol manipulation to explain cognition, \emph{emergent models} focus on the abilities that emerge from the relationship between autonomous systems and their environment. According to McClelland et al. \cite{mcclelland2010letting}, the emergent approach is based on the idea that operations of sub-cognitive processes result in behavior; thus, aims at modelling cognitive processes at a sub-symbolic level. Since our focus is on the emergent models, we structured our survey following \cite{mcclelland2010letting}. For cognitive models, one may refer to \cite{anamalamudi2014computational} and \cite{komosinski2015time}.

%  to include neurocomputational models of time perception.
% The embodiment is not crucial for cognitive models, whereas, for emergent models, the embodiment is a must. Another explanation of emergent approach can be found in the cognitive science literature.

In addition to the emergence, embodiment is another property of computational models. We accepted that embodied models are the models forming their experience through \emph{``sensory and bodily interaction with their environment''} \cite[p. 303]{mainzer2009embodied} and disembodied models are the models that do not focus on experience formation. Relying on this definition, we grouped emergent models as \emph{embodied emergent models} and \emph{disembodied emergent models}. Recall that time perception models are grouped into two classes in the literature, namely dedicated and intrinsic models. This gives us a chance to further categorize disembodied emergent models into two kinds as \emph{disembodied emergent dedicated models} and \emph{disembodied emergent intrinsic models}. The employed categorization is given in Fig.~\ref{fig:time_perception_models} and a summary of computational and robotic models is given in Table \ref{tab:com_rob_tim_real}.

\begin{table*}[!t]
\footnotesize
\centering
\caption{The Emergent Computational and Robotic Models of Time Perception}
\vspace{-12pt}
\begin{tabular}{M{7mm}M{40mm}M{30mm}M{5mm}llllM{40mm}}

\label{tab:com_rob_tim_real}\\
\hline

Cat &
  Name &
  Mechanism &
  SP &
  ST &
  MT &
  PT &
  RT &
  Comment
\\
DEDM &
  Perception-based model \cite{roseboom2019activity} &
  Counting salient change &
  \tick &
  \tick &
  - &
  \tick &
  - &
  accounts the effect of perceptual content. \\
 &
  MOMs: BF \cite{miall1989storage} &
  Oscillations &
  \cross &
  \tick &
  - &
  \tick &
  - &
  the first multiple oscillator model. \\
 &
  MOMs: SBF \cite{buhusi2013time} &
  Tracking oscillations &
  \tick &
  \tick &
  - &
  \tick &
  - &
  the first perceptron-based realization of multiple-oscillator models. \\
 &
  MDMs: GAMIT-net \cite{addyman2014gamit} &
  Exploiting memory decay process &
  \tick &
  \tick &
  - &
  \tick &
  \tick &
   shows wide range of abilities. \\
 &
  EMs: Neuro-evolutionary optimization \cite{maniadakis2016and} &
  Universal timing module &
  \cross &
  \tick &
  - &
  \tick &
  - &
  the first telling when an event happened. \\ \hline
DEIM &
  Synfire chain model \cite{hass2008neurocomputational} &
  Synchronous firing of chains &
  \tick &
  \tick &
  - &
  \tick &
  - &
  considers millisecond-based interval timing. \\
 &
  Positive-feedback model \cite{gavornik2009learning} &
  Reward modulated plasticity &
  - &
  \tick &
  - &
  \tick &
  - &
  does not assume a special neuron type. \\
 &
  {\color{black}Delay network model \cite{de2019flexible}} &
  {\color{black}Representing input history} &
  {\color{black}{\tick}} &
  {\color{black}{\tick}} &
  {\color{black}{-}} &
  {\color{black}{\tick}} &
  {\color{black}{-}} &
  {\color{black}extendable and captures violations of scalar property.} \\
 &
  State-dependent network and population clock models \cite{karmarkar2007timing, hardy2018encoding} &
  State-dependent changes in neural properties &
  \tick &
  \tick &
  \tick &
  \tick &
  - &
  state-dependent neural properties are exploited. \\ \hline
EEM &
  MDMs: Developmental robotics model \cite{addyman2011learning} &
  Exploiting memory decay process &
  \tick &
  \tick &
  - &
  \tick &
  - &
  the first embodied model. \\
 &
  EMs: Duration comparison \cite{maniadakis2012experiencing} &
  Inverse ramping activity &
  - &
  \tick &
  - &
  \tick &
  - &
  a self-organizing system. \\
 &
  EMs: Duration comparison and production \cite{maniadakis2014robotic} &
  Clock-like mechanism counting imperfect oscillations &
  - &
  \tick &
  \tick &
  \tick &
  - &
  integrates dedicated and intrinsic representations. \\
 &
  EMs: Duration comparison, production and categorization \cite{maniadakis2015integrated} &
  Clock-like mechanism counting imperfect oscillations &
  - &
  \tick &
  \tick &
  \tick &
  - &
  integrates dedicated and intrinsic representations. \\
&
RLs: Feedforward agent \cite{deverett2019interval} &
Autostigmergic behavior &
\cross &
- &
\tick &
\tick &
- &
uses environment to store temporal information. \\
 &
  RLs: Recurrent agent \cite{deverett2019interval} &
  Ramping/inverse ramping activity &
  - &
  - &
  \tick &
  \tick &
  - &
  RL agent can process temporal information. \\
&
 {\color{black}{RLs: twofold model \cite{lourencco2019teaching}}} &
  {\color{black}{sensory information and RL}} &
  {\color{black}{\tick}} &
  {\color{black}{\tick}} &
  {\color{black}{-}} &
  {\color{black}{\tick}} &
  {\color{black}{-}} &
  {\color{black}{a biologically plausible approach.}} \\
 &
  DNFs: mobile robot \cite{duran2017learning} &
  Accumulation of memory trace &
  - &
  - &
  \tick &
  \tick &
  - &
  a realization of an intrinsic model in a mobile robot. \\
 &
  {\color{black}{DNFs: Temporal order \cite{ferreira2011dynamic, ferreira2014learning, wojtak2015learning}}} &
  {\color{black}{Decaying neural activation and threshold dynamics}} &
  {\color{black}{-}} &
  {\color{black}{-}} &
  {\color{black}{\tick}} &
  {\color{black}{\tick}} &
  {\color{black}{-}} &
  {\color{black}{consider the temporal order between events.}} \\
 &
  {\color{black}{DNFs: Temporal order and interval timing \cite{wojtak2019neural, ferreira2020rapid}}} &
  {\color{black}{Decaying neural activation, threshold dynamics and ramping activity}} &
  {\color{black}{\cross}} &
  {\color{black}{\tick}} &
  {\color{black}{\tick}} &
  {\color{black}{\tick}} &
  {\color{black}{-}} &
  {\color{black}{consider the temporal order between and duration of events.}} \\
 \hline
ORM &
  Temporal prediction model  \cite{hourdakis2018robust} &
  Learning temporal features of actions &
  - &
  \tick &
  - &
  \tick &
  - &
  one of the first studies in the field. \\
 &
  Action learning model \cite{koskinopoulou2018learning} &
  Learning spatio-temporal features of actions &
  - &
  - &
  \tick &
  \tick &
  - &
  one of the first studies in the field.\\
\hline
\begin{minipage}{17cm}%
    \vspace*{0.2cm}
    \textit{Note:} Cat: Categories, DEDM: Disembodied Emergent Dedicated Models, DEIM: Disembodied, Emergent Intrinsic Models, EEM: Embodied Emergent Models, ORM: Other Robotic Models, SP: Scalar Property, ST: Sensory Timing, MT: Motor Timing, PT: Prospective Timing, RT: Retrospective Timing, MOMs: Multiple Oscillator Models, MDMs: Memory Decay Models, EMs: Evolutionary Models, RLs: RL Models, DNFs: Dynamic Neural Field-Based Models. \tick: model shows the property or ability. \cross: model does not show the property or ability. -: model does not aim for capturing the property or ability%
  \end{minipage}%
\end{tabular}
\end{table*}

\subsection{Disembodied Emergent Models} \label{disembodied_intrinsic}

% According to Ziemke \cite{ziemke2003s, ziemke2004embodied}, a system can be embodied \emph{physically, biologically, socially and historically}.
%In this section, we will discuss disembodied and emergent models of time perception (Fig.~\ref{fig:time_perception_models}).
%These models are \emph{neurocomputational} and \emph{neural network-based models}.

% Emergent models assuming specialized functions are considered as dedicated models, whereas those focusing on the temporal processing abilities of neurons are considered as intrinsic models.

\subsubsection{Disembodied emergent dedicated models} 

Dedicated models assume that temporal information processing depends on specialized systems or functions in the brain \cite{ivry2008dedicated}. 

% Two types of disembodied emergent dedicated models can be defined depending on how the internal clock transforms physical time into subjective time. These models are pacemaker-accumulator models and multiple-oscillator models. They assume different physical realizations that result in a clock-like function.

% Since the models are neural-network-based, they can be considered as emergent models. Since models assume a dedicated mechanism for time perception, they can be regarded as dedicated models.

\begin{paragraph}{Pacemaker-accumulator models} These models are currently the most prevalent models in the literature \cite{addyman2016computational, simen2013timescale}. They assume that a pacemaker generates pulses and an accumulator collects them. This idea was realized by cognitive architectures \cite{addyman2016computational, pape2008model, taatgen2007integrated, taatgen2008time} and mathematical models \cite{gibbon1992ubiquity, killeen2000propagation}. Since the pacemaker-accumulator model and the internal clock theory share similar assumptions, the same disadvantage applies to both. In the literature, the number of emergent realizations of the pacemaker-accumulator model is scarce. An emergent version of the pacemaker-accumulator model \cite{roseboom2019activity} will be discussed next. \end{paragraph}

% In a mathematical model, Gibbon assumed that the scalar property arises from the noisy nature in the \emph{comparison} of observed and remembered duration \cite{gibbon1992ubiquity}. Recall that the scalar property is given in Figure \ref{fig:scalar_property}. Apart from the mathematical model developed by Gibbon, other variants of the pacemaker-accumulator model were proposed in the literature. For example, Killeen and Taylor \cite{} assumed that some pulses that are sent to the accumulator from the pacemaker are not accumulated. The noise in the accumulation process makes the mean and the standard deviation of accumulated pulses linearly proportional.
 
%  Functions used in the model to simulate the human performance were hand-crafted such that they form similar structures to phases defined in the internal clock theory. According to Simen et al. \cite{simen2013timescale}, major critiques to pacemaker-accumulator models are due to their lack of correspondence to behavioral data and neural evidence.

\paragraph{Perception-based model}

Pacemaker-accumulator models do not assume a relationship between sensory information and time perception, even though temporal information is acquired through sensory modalities. Roseboom et al. \cite{roseboom2019activity} proposed that counting salient \emph{visual change} is the primary mechanism of time perception. To test this idea, they adopted a transfer learning approach by using a deep image classifier \cite{krizhevsky2012imagenet}. The model calculated the Euclidean distance between the activation values of layers formed for each frame for detecting changes and accumulated these changes when they exceed a dynamically set threshold. The accumulated changes were used to predict subjective time estimation. The model can be considered as an emergent realization of the pacemaker-accumulator model given that the accumulated salient changes may be thought as ticks of a pacemaker. Roseboom et al. \cite{roseboom2019activity} observed that the model performed well in duration estimation tasks. Although the model has the ability of prospective sensory timing and approximates scalar property, it does not address retrospective and motor timing abilities. Being vision-based, findings of the model might be seen as limited, but the idea itself is scalable to other modalities, which implies a possible research direction for interval timing. Recently, Fountas et al. \cite{fountas2020predictive} extended \cite{roseboom2019activity} by developing a hierarchical Bayesian model of episodic memory and time perception to capture the effects of attention, cognitive load and scene type on time perception. They integrated semantic and episodic memory into time perception in a \emph{predictive processing model} that was composed of bottom-up and top-down processes. 

% However, the model is not emergent as it uses hierarchical Bayesian modeling to estimate salient changes and exploits experimental data for parameter fitting. 

\begin{paragraph}{Multiple-oscillator models} These models assume that internal clock functions are realized by the areas emitting oscillations in different frequencies in the brain \cite{matell2004cortico}. They are categorized into \emph{Beat Frequency (BF)} and \emph{Striatal Beat Frequency (SBF)} models \cite{buhusi2013time, matell2004cortico, miall1989storage}. The BF model \cite{miall1989storage} assumes more than one oscillator. Each oscillator oscillates in a different frequency and resets into the same level when a new stimulus arrives. Phases of oscillators estimate the duration of the stimulus. However, the BF fails to satisfy the scalar property. Buhusi and Oprisan \cite{buhusi2013time} improved the SBF model to achieve the scalar property. The model was a perceptron showing activation to oscillatory inputs, which were stored in long-term memory in reinforced trials (as in FI). When the current activation in working memory was similar to the activation stored in the long-term memory, output neurons showed activation. Therefore, the system learned when the reward is delivered based on the activation patterns of a perceptron. According to researchers \cite{buhusi2013time}, noise applied to the parameters of the model led to the emergence of scalar property.

\end{paragraph}

\paragraph{Memory decay models} They hold the view that interval timing ability is grounded on \emph{memory decay processes}. According to the decay theory, as time passes, the information in short-term memory fades away and forgetting occurs \cite{lewandowsky2009no}. Addyman and Mareschal \cite{addyman2014gamit} developed an interval timing model called GAMIT-net based on this phenomenon. It is a recurrent neural network (RNN) model that receives a Gaussian distribution simulating memory decay processes as input and estimates time. GAMIT-net was trained for retrospective and prospective timing. For retrospective timing, to simulate inattentiveness, the model was trained on initial and final points of each event, whereas for prospective timing, it was trained occasionally. As a result, the model captures the prospective and retrospective timing performances, mimics the scalar property because of the noisy nature of the simulated memory decay, and shows the effects of working memory load and attention on interval timing. Despite its explanatory capabilities, it is unclear whether the memory decay process is responsible for forgetting \cite{lewandowsky2009no}.

%  Since GAMIT-net proposes a specialized mechanism for time perception, we considered this as a dedicated timing model.
% is time rather than the interference of new information \cite{lewandowsky2009no}. 

\subsubsection{Disembodied emergent intrinsic models} \label{deim} These models are disembodied emergent models assume that neurons can process temporal information without a specialized cognitive mechanism. They seem to be limited to the millisecond scale (10 and 100 ms) \cite{block2014time, ivry2008dedicated, paton2018neural}. 

\begin{paragraph}{Synfire chain model} Synfire chain model was developed by Ha{\ss} et al. \cite{hass2008neurocomputational}, who proposed that chained neuron groups fire synchronously to represent temporal information. Each chain is composed of neurons activating one another and the temporal estimation errors are determined by the differences between lengths of these chains. According to the model, timing is achieved by the estimations of different chains. The model tracks time prospectively and shows the scalar property due to the cumulative error in the combination of temporal estimations. Being extendable to motor timing, the model, however, fails to address retrospective timing because it needs a starting point to track time.
\end{paragraph}

% result in different estimation capabilities of observed time due to the accumulated error correlating with the length of chain. 
 
\begin{paragraph}{Positive-feedback model} This model tries to explain how mice learn a specific duration in FI by exploiting the temporal correlation between a reward-based visual stimulus and the elapsed time \cite{gavornik2009learning}. This correlation is utilized by a RNN that can exhibit \emph{reward-dependent plasticity}. It was observed that, after the training, the model shows a sustained activity until the reward is received. The model is based on prospective timing and not capable of retrospective timing.
\end{paragraph}

{\color{black} \begin{paragraph}{Delay network model} Recently, making use of Neural Engineering Framework (NEF), de Jong et al. \cite{de2019flexible} adopted a novel approach, which captured scalar property and its violations. Additionally, they assumed a connection between the scalar property and a neurological phenomenon called neural scaling \cite{wang2018flexible}, which is the property of neurons whose firing distributions stretch or compress according to the temporal length of a given stimulus. This is analogous to scalar property, which is the stretch and compress of response distributions dependent on stimulus duration. To test this hypothesis, they utilized the delay network \cite{voelker2018improving}, which is an RNN maintaining a temporal memory by compressing its input history. They trained the network to produce several durations with different values of memory size, and showed that the model fits well with the empirical data of interval timing and neural scaling. This way, the model also captured violations of scalar property in short intervals ($<$ 500 ms) observed by \cite{fetterman1992time}. According to the authors, the model can be extended to a wide range of timing abilities, integrated with other cognitive models, and utilized to investigate how temporal learning occurs. 
\end{paragraph}}

\begin{paragraph}{State-dependent network and population clock models} These models assume that temporal information processing is a result of recurrently connected neural populations. While the state-dependent network model was developed for sensory timing \cite{karmarkar2007timing}, the population clock model was developed for motor timing abilities \cite{buonomano2011population}. According to the state-dependent network model, neural populations code temporal information via their \emph{synaptic, cellular, and structural properties}. Event-related stimuli lead to \emph{short-term plasticity}; in other words, change in the hidden state of the neural population and activation patterns \cite{hardy2016neurocomputational}. The change in hidden state can be used to detect the duration of an event. The model, therefore, transforms temporal information into a spatial pattern \cite{karmarkar2007timing}. To illustrate the process, imagine skipping a stone where each bounce leads to a change in the water, as a consequence, patterns in the water may reflect the properties of the stone. Several studies were conducted to examine the explanatory capabilities of state-dependent network models \cite{buonomano1995temporal, buonomano2009state, hardy2018encoding, perez2018synaptic}. In their seminal work, Karmarkar and Buonomano \cite{karmarkar2007timing} developed a neurocomputational model, which was composed of recurrently connected 400 excitatory and 100 inhibitory neurons. Each neuron could show short-term synaptic plasticity and inhibitory postsynaptic potential. In their simulation, Karmarkar and Buonomano \cite{karmarkar2007timing} visualized dynamics of two networks, one of which received one auditory stimulus and the other received two auditory stimuli 100 ms apart. They observed that network dynamics, encoding stimulus history, can be represented as a spatial code, which then can be utilized to receive temporal information.

According to the population clock model \cite{buonomano2011population}, there are two systems that work together: One of these systems is the population clock composed of neurons exhibiting activation patterns in response to an incoming stimulus; whereas the other is the read-out neuron reading these activation patterns. Neural activations in the population clock for the time point at which behavior is executed are higher than other time points, which can be detected by the read-out neuron \cite{buonomano2011population}. Hardy and Buonomano \cite{hardy2018encoding} developed a neurocomputational model to test whether sequential activation of neurons can encode temporal features of behaviors. To achieve this, they trained an RNN, formed by excitatory and inhibitory connections between neurons, for generating a 5-sec target trajectory. It turned out that the model successfully produces the given trajectory and the performance obeys the Weber's law \cite{hardy2018encoding}. Both models can track time prospectively.
\end{paragraph}

%These models above are disembodied because, in the simplest case, they do not consider an agent who senses and acts. In the next subsection, we will cover embodied emergent models of time perception.

\subsection{Embodied Emergent Models}
The literature indicates a strong relationship between embodiment and temporal experience. For example, recent research showed that bodily and emotional states affect time perception \cite{wittmann2013inner}. Moreover, some researchers consider that temporal representation is formed via bodily and emotional states \cite{craig2009you, di2018feel}. According to Craig et al. \cite{craig2009you}, the posterior side of the insular cortex integrates bodily states and motivational factors to form temporal representations. Also, supplementary motor area is responsible from the relationship between behavior and temporal representations \cite{coull2016act, merchant2016motor}. Addyman et al. \cite{addyman2017embodiment}, addressed the possibility that interval timing depends on the development of the motor system. In short, the accumulated evidence suggests that the embodiment is important for time perception (for a general review, see \cite{kranjec2010temporal}). In this section, we will explain the embodied emergent models of time perception.

We categorize studies in the literature into four groups, namely,  \emph{memory decay processes} based \cite{addyman2011learning}, \emph{evolutionary optimization} based \cite{maniadakis2009explorations,maniadakis2011time, maniadakis2012experiencing, maniadakis2015integrated, maniadakis2016and}, \emph{reinforcement learning} based \cite{deverett2019interval} {\color{black}Lourenço et al. \cite{lourencco2019teaching}}, and \emph{dynamic-neural field theory based} \cite{duran2017learning}, {\color{black}\cite{ferreira2011dynamic, wojtak2015learning, wojtak2017towards, wojtak2019neural, ferreira2020rapid}} models.

\subsubsection{Memory decay models}
Addyman et al. \cite{addyman2011learning} developed the first embodied model of interval timing for the developmental emergence of interval timing. They explored the possibility that memory decay can be converted into duration information. For example, a baby trying to reach a toy has a memory of the toy decaying over time, and at the same time, a reaching behavior. The association between the duration of the motor behavior and the memory decay is formed during the development  and re-used for interval timing. Addyman et al. \cite{addyman2011learning} tested this hypothesis using an RNN model, where the input is simulated via visual and auditory information derived from a Gaussian distribution, and the output is a one-hot encoded arm movement. The model is capable of prospective sensory timing and shows the scalar property, yet does not aim to explain retrospective and motor timing. In a later study, Addyman et al. \cite{addyman2014gamit} demonstrated that the memory decay process is expandable to retrospective timing. The model receives action information as one-hot encoded vector and sensory information as a fading Gaussian distribution. These simulation-based decisions might make it difficult to generalize findings of the model to real life. It is worth emphasizing that the model, in principle, builds a mapping between the length of the performed action and its duration, which might relate this model to hypotheses trying to explain the relationship between different magnitudes \cite{walsh2003theory}. Taking simple simulation-based decisions, Addyman et al. \cite{addyman2011learning} tested the role of memory-decay in interval timing. Time perception mechanisms can be investigated without simplifying the simulation and assuming a theory. Maniadakis et al. \cite{maniadakis2012experiencing, maniadakis2014robotic, maniadakis2015integrated} utilized evolutionary optimization as a method to develop models that have little or no assumption in order to investigate possible time perception mechanisms.

\begin{figure}[t]
\centering
  \includegraphics[width=\columnwidth]{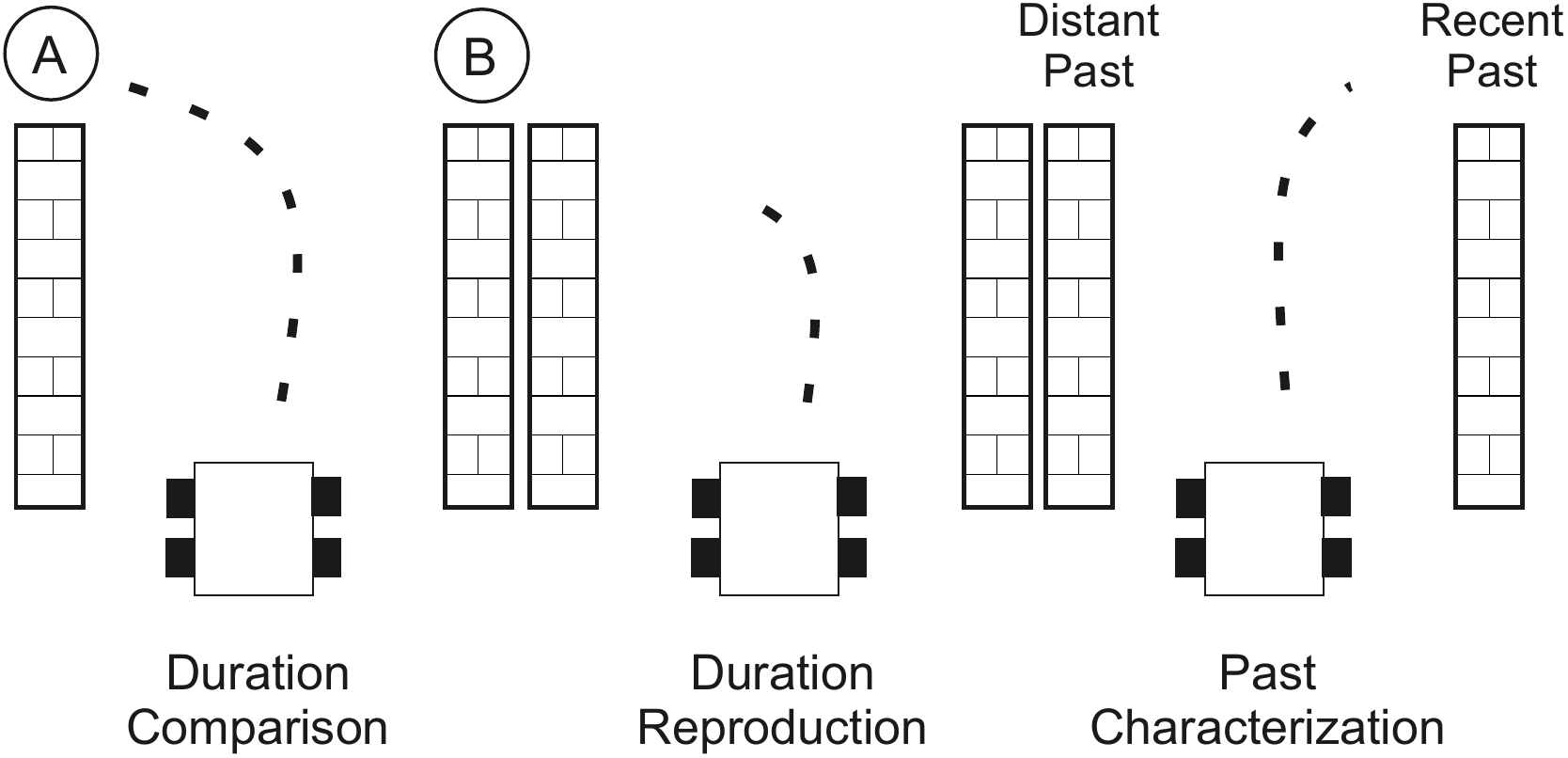}
\caption{{\color{black}The tasks in \cite{maniadakis2012experiencing, maniadakis2014robotic, maniadakis2015integrated}. The mobile robot should go right or left to indicate its decision in the duration comparison and past characterization tasks. For the duration reproduction task, it should continue to move for the time length given. }}
%The figure is adapted from \cite{maniadakis2012experiencing, maniadakis2014robotic, maniadakis2015integrated}.}}
\label{fig:evoptimization}
\end{figure}

\subsubsection{Evolutionary models}
Maniadakis et al. \cite{maniadakis2012experiencing, maniadakis2014robotic} developed a \emph{continuous-time RNN} for sensory and motor timing using an evolutionary optimization procedure. In these studies, a mobile robot and a simulation environment were used (Fig.~\ref{fig:evoptimization}). While \cite{maniadakis2012experiencing} trained the model only with a sensory timing task, namely \emph{duration comparison}, \cite{maniadakis2014robotic} extended the same idea to motor timing task, namely \emph{duration reproduction} (Fig.~\ref{fig:temporal_abilities}). For duration comparison, the mobile robot had to decide which stimulus is longer by turning left or right at the end of the corridor. For duration reproduction, the mobile robot had to continue its movement for the overall duration of the target stimulus. The model trained for only duration comparison exhibited an inverse ramping activity and its dynamics were not similar to a clock \cite{maniadakis2012experiencing, gibbon1977scalar, gibbon1984scalar, treisman1963temporal}. In a subsequent study \cite{maniadakis2014robotic}, the model was also trained for duration reproduction. It revealed a ramping activity and imperfect oscillatory activations similar to ticks of a clock, conforming to the assumptions of multiple-oscillator models. Furthermore, \cite{maniadakis2014robotic} investigated the role of embodiment and observed that neurons responsible for action execution were also used for timing, confirming the assumptions of intrinsic models \cite{ivry2008dedicated}. Whereas the model developed by \cite{maniadakis2012experiencing, maniadakis2014robotic} is successful at prospective sensory timing, models of \cite{maniadakis2014robotic} are successful at prospective motor timing.

Maniadakis and Trahanias \cite{maniadakis2015integrated} addressed a \emph{past characterization} skill (Fig.~\ref{fig:evoptimization}), where the model decides whether an event occurs in the near or distant past. Related to the memory processes, past characterization shares similarities with retrospective timing. After training, the network exploited both a clock-like mechanism counting oscillatory activations and their amplitudes. This is contrary to the assumptions of pacemaker-accumulator models which assume that a tick/pace corresponds to one static temporal unit \cite{gibbon1984scalar}. The clock developed by this system had both a \emph{count-up mechanism} working during duration estimation and a \emph{count-down mechanism} working during duration reproduction. Moreover, neurons serving for ordinary tasks were also utilized for representing time. Since the model showed properties of both intrinsic and dedicated models, \cite{maniadakis2015integrated} concluded that both models might be implemented into the artificial brains.

In a further study, Maniadakis and Trahanias \cite{maniadakis2016and} took one step further by developing a model that can assess both \emph{when} and \emph{how long} an event occurred. They adopted an incremental neuro-evolutionary optimization approach. In the first phase, they trained models to assess how long an event takes place and when an event occurs. The system received oscillations in four different frequencies as input, conforming to the assumptions of the SBF model \cite{buhusi2013time} (recall that SBF is a multiple-oscillator model). Receiving the oscillatory signals, a universal time source generated a composite time representation. After then, the generated time representation coupled with an event label was sent to \emph{working memory}. Consequently, in the second phase, the model was trained for tracking when an event occurred. The model developed in the first phase was a prospective sensory timing model that tracks time for more than one event, which is an improvement over interval timing models. However, the model failed to mimic the scalar property. Considering the second phase, the model can be thought of as a retrospective sensory timing model that continuously tracks the passage of time and stores temporal information for more than one event. The formation of a universal timing module implies both the amodality of temporal representations and the validity of dedicated models.

% They do not aim to capture the relationship between the perceptual content and time perception. However, it is widely accepted that time perception is multi-modal \cite{bausenhart2014multimodal, vroomen2010perception} and affected by perceptual content \cite{roseboom2019activity}. 

{\color{black}{As another unbiased way of exploring time perception mechanisms, we will next investigate reward-based end-to-end models.}}

\begin{figure}[t]
\centering
  \includegraphics[width=\columnwidth]{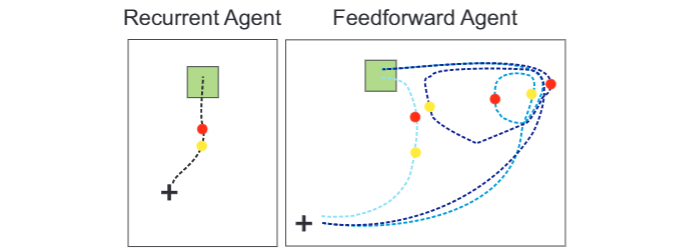}
\caption{{\color{black}The behaviors of feedforward and recurrent agents in solving duration reproduction task \cite{deverett2019interval}. The cross, yellow and red dots show the initial position, ready and set cues, respectively. After the ready cue, the required interval is given. After the set cue, the interval is reproduced by the agent. The dashed lines show possible solutions. The recurrent agent solves the task by counting the temporal interval. The feedforward agent exhibits spiral-like behaviors, where the width of spiral depends on the estimated interval.}}
\label{fig:deeplearningagent}
\end{figure}

\subsubsection{RL based models}

Deverett et al. \cite{deverett2019interval} investigated interval timing in \emph{deep RL} in a duration reproduction task, similar to the task employed by \cite{maniadakis2014robotic} (Fig.~\ref{fig:temporal_abilities}). The agent was represented with a point which acts on and gets reward from a two-dimensional simulation environment informing agent about the current phase of the trial. By using two types of networks (feedforward or long-short term memory), the authors generated two types of agents. Although the feedforward agent learned slower and had poorer generalization, both each agent could learn the task. Investigating the hidden layer of LSTM of the recurrent agent with principal component analysis (PCA), Deverett et al. \cite{deverett2019interval} found out that the system exhibits an accumulating pattern until the duration reproduction phase starts and a reducing pattern while the model produces the interval. This activation is similar to the counting mechanism observed by \cite{maniadakis2015integrated, duran2017learning}. On the other hand, a systematic pattern was not found in the activations of the feedforward agent. To reveal how the agent, who is not capable of memorizing, can solve the task, researchers looked over trajectories of agents while solving the task. As seen in Fig.~\ref{fig:deeplearningagent}, the feedforward agent developed an \emph{autostigmergic behavior} to use the environment as a temporal information source. {\color{black}Its success can be accounted by BeT and LeT models (Fig.~\ref{fig:BETLET}, Section II-B3).}

Deverett et al. \cite{deverett2019interval} showed the importance of recurrent information processing for developing a clock-like mechanism and a possible behavioral strategy employed to exploit the environment as a temporal information source. Whether agents trained by evolutionary optimization develop similar behavioral strategies for timing might provide evidence for BeT and LeT (Section II-B3). We believe that same RL algorithm might be applied to other timing tasks for exploring time perception mechanisms. Summarizing capabilities of models, we can say that both models \cite{deverett2019interval} are capable of prospective motor timing and the recurrent agent's temporal estimations obey scalar property. 

{\color{black}Instead of being used for developing agents end-to-end manner, RL can be used to simulate dopaminergic system controlling agents' actions, as was done by Lourenço et al. \cite{lourencco2019teaching}. Taking inspiration from time perception literature, researchers identified two different mechanisms, namely external and internal timing. While the former was assumed to be an innate ability of organisms, the latter was thought as a simulation of the dopaminergic system undertaking time-related tasks. The external timing mechanism was modeled by Gaussian processes to derive temporal information from sensory observation, whereas the internal timing mechanism was modeled by temporal difference learning. The model was tested with a duration discrimination task, which requires deciding whether a given interval is longer or shorter from another interval. As a result, researchers demonstrated that the model was successful at approximating the given interval, the variance in estimation increased with the interval length (approximately scalar), and classification uncertainty (whether given interval is short or long) increased at boundaries.}

\subsubsection{Dynamic neural field (DNF) based models}

DNF is a mathematical formulation about how neural populations work. Assuming the principles of DNF, Duran and Sandamirskaya \cite{duran2017learning} developed a model that can learn and represent the duration of actions and tested their model in a mobile robot. The mobile robot had to navigate between locations while avoiding objects. The model was based on elementary behaviors that represent the relationship between neural states and actions. {\color{black}Each elementary behavior had three DNFs: intention DNF signifying onset of action, condition of satisfaction DNF (CoS) controlling whether is action is completed, and dissatisfaction DNF (CoD) checking whether the action should be aborted in case the goal could not be achieved. A node called \emph{t} regulated the competition between CoS and CoD. } In the earlier trials of training, \emph{t} gave an advantage to CoS because the duration of the action is unknown, whereas in the later trials, \emph{t} gave an advantage to CoD because the action (therefore temporal dynamics of the action) was learned. By this way, consistent memories were accumulated and controlled the agent's behavior. Researchers showed that the model could represent, store, and update temporal information; moreover, it could detect anomalies based on unexpected temporal differences. The model instantiates a state-dependent network because how time is represented depends on the current state of the network. As with the majority of models discussed in this section \cite{deverett2019interval, maniadakis2015integrated, maniadakis2016and}, this model also showed a ramping activity. Findings of this study show the possibility of realizing intrinsic models of time perception in robots to extend their capabilities from milliseconds to seconds.

{\color{black}A set of preliminary studies employing the principles of DNF to endow robotic systems with timing ability was conducted by Ferreira, Wojtak, and their colleagues \cite{ferreira2011dynamic, wojtak2015learning, wojtak2017towards, wojtak2019neural, ferreira2020rapid}. In general, their primary objective was to develop robots, being capable of precise timing for smooth human-robot interaction. In their initial work, Ferreira et al. \cite{ferreira2011dynamic} coded ordinal features of behaviors in a sequence by decaying neural activations and next \cite{ferreira2014learning} realized the model in a robot and demonstrated its capability in a music playing task. Subsequent research integrated sensory and proprioceptive feedback to the model \cite{wojtak2015learning} to tune ordinal and temporal representations of behaviors. Following research scaled the model to an assembly task, in which the robot had to learn steps via observation \cite{wojtak2017towards}.

Wojtak et al. \cite{wojtak2019neural} developed a DNF-based computational model for duration estimation and reproduction. The temporal information was represented by the amplitude of the corresponding self-stabilized bump and researchers observed ramping characteristics in the network dynamics. Recently, Ferreira et al. \cite{ferreira2020rapid} broadened the scope of their computational model \cite{ferreira2014learning, wojtak2017towards} to interval timing, extending the previous approach to learning both event orders and durations. The model can execute a sequence of behaviors in different execution speeds while preserving relative timing. However, temporal estimations do not approximate scalar property.}

\subsubsection{Other robotic models}

For learning the temporal dynamics of actions, Hourdakis and Trahanias \cite{hourdakis2018robust} developed a computational model composed of two components: task progress and control. The former is responsible for detecting how much of a given task is completed, whereas the latter tracks the primitive motions of actions. Koskinopoulou et al. \cite{koskinopoulou2018learning} extended the learning from demonstration framework for learning both spatial and temporal dynamics of an action. This allows executing actions at variable speeds and forming temporal plans.

\section{Discussion and Conclusion} \label{dc}

One of the most critical discussions in the literature is whether time is processed and represented by intrinsic or dedicated systems. Considering the embodied models of time perception, we could list several mechanisms: oscillatory activations that are counted by a clock-like mechanism \cite{maniadakis2014robotic, maniadakis2015integrated} and ramping activity in neural activations \cite{duran2017learning, deverett2019interval, maniadakis2014robotic}. It seems that embodied models tend to validate dedicated models of time perception proposing an internal clock tracking time. However, the clock proposed in these models deviates from the original dedicated models. For example, Maniadakis et al. \cite{maniadakis2012experiencing, maniadakis2015integrated} reported a dynamic temporal pace that is simulated by oscillatory activations. Deverett et al. \cite{deverett2019interval} observed a clock-like activation after PCA, which shows the possibility of a distributed version of dedicated models. On the other hand, using a universal time source module \cite{maniadakis2016and}, which shows a wide range of temporal abilities, might be a means to assess dedicated representations of time. The use of a universal time source also implies that temporal representations are amodal. Overall, this review demonstrates that the development and exploration of artificial agents perceiving time reveals insights on how time is represented and processed in natural cognitive systems.

It is generally accepted that intrinsic models of time perception are limited to millisecond range \cite{ivry2008dedicated}. State-dependent networks, which are categorized as intrinsic models, face the same challenge. Duran and Sandamirskaya \cite{duran2017learning} implemented a model conforming to the principles of state-dependent networks in a robot that could track time successfully. This implies the possibility of enhancing the capabilities of intrinsic models to seconds through the medium of embodiment.

It was noted that time perception cannot be easily achieved by expanding the capabilities of intrinsic models to seconds. As already stated, time perception has at least four timescales, namely circadian, second, millisecond, and microsecond timing \cite{buhusi2005makes}. Considering the role of the time in forming experiences, we can add hours, days, weeks, months, and years to this list. Furthermore, we can double the list by considering the past and the future, which surely will make the problem more complicated. Developing artificial systems that perceive the past and present requires combining time perception with other cognitive abilities and mechanisms like working memory, as in \cite{maniadakis2016and, ferreira2011dynamic}, long-term memory, and attention. To succeed in this endeavor, RL and DNF can be plausible approaches for integration between cognitive abilities. 

In addition to the importance of multiple time-scales in cognitive life, Table \ref{tab:com_rob_tim_real} shows that retrospective timing, learning temporal features of the environment implicitly, is largely unexplored by emergent and embodied models. 
{\color{black}Besides retrospective timing, studies on how animals learn complex temporal dynamics of action sequences \cite{wojtak2019neural,ferreira2020rapid,hourdakis2018robust,koskinopoulou2018learning} are promising. It should be emphasized that the majority of artificial agents investigated in this review are unifunctional. For instance, the models developed by Roseboom et al. \cite{roseboom2019activity}, Addyman et al. \cite{addyman2011learning}, and Maniadakis et al. \cite{maniadakis2016and} are capable of sensory timing, whereas the models developed by Deverett et al. \cite{deverett2019interval} and Duran et al. \cite{duran2017learning} are capable of motor timing. It is highly probable that similar algorithms can be extended to accomplish a wider range of time perception tasks and abilities to get insights about time perception mechanisms.}

It is necessary to note that temporal abilities discussed in this review are highly limited and only represent a small proportion of the field. For example, we did not include the verbal estimation of time \cite{block2018prospective} and processing temporal information of sequences \cite{hardy2016neurocomputational}, partly because of the sparsity of emergent models in these aspects. We also could not spare enough space to discuss other cognitive abilities such as mental time travel, reasoning about the future and time-dependent organizations of memory. It is exciting that further studies can investigate new abilities in artificial systems and gain insights into how natural systems can solve these problems. 

In their inspirational work, Roseboom et al. \cite{roseboom2019activity} suggested that tracking salient change in the perceptual content might be a mechanism of interval timing and reported that it is possible to ground interval timing on sensory information. {\color{black}A similar, twofold learning approach to time perception was employed in the model of Lourenço et al. \cite{lourencco2019teaching}, who extracted time from sensory information. Whether the very same idea can be extended to other time perception abilities and robotic agents operating in the real world is an appealing question to discuss.}
 
As essential parts of cognitive life, the perception of other magnitudes seems to have a evolutionary and developmental relationship to the perception of time \cite{walsh2003theory, cantlon2012math}. Perhaps, artificial agents trained incrementally or holistically for using different magnitudes can be assessed for possible overlapping mechanisms. The evolutionary optimization approach proposed by Maniadakis et al. \cite{maniadakis2012experiencing, maniadakis2014robotic} 
can be utilized for a minimally-biased exploration of a common magnitude system. It is also possible that one can build a bridge between sensory-motor decision variables \cite{walsh2003theory} and the use of space to estimate intervals based on decaying memory trace over time \cite{addyman2011learning, addyman2017embodiment}. This might connect time, space, and number based on actions resulting in embodied timing models capable of using magnitudes for action selection and control.

The scalar property shows exciting challenges to computational and robotic models of time perception. 
From an application point of view, as long as temporal estimations are accurate enough, addressing the scalar property in artificial systems might not be necessary. On the other hand, from a scientific point of view, as temporal estimations imposed by scalar property are shared among animals \cite{buhusi2005makes, ferrara1997changing, lejeune2006scalar, matell2004cortico, malapani2002scalar, wearden1997scalar}, addressing it in artificial systems might ease the generalization of the results to biological systems.

This paper discussed time perception through the lens of a wide range of disciplines. Considering the role of time in natural systems, we consider time perception as a present challenge to be met by artificial intelligence and a possible way to develop robust and adaptive systems. We also believe that developing computational and robotic systems reveal significant insights into how biological time perception emerges.

% \ifCLASSOPTIONcaptionsoff
%   \newpage
% \fi





% \end{thebibliography}

\bibliographystyle{IEEEtran}
\bibliography{IEEEabrv, main}

\end{document}